\documentclass[conference]{IEEEtran}
\IEEEoverridecommandlockouts

\usepackage{cite}
\usepackage{amsmath,amssymb,amsfonts}
\usepackage{algorithmic}
\usepackage{graphicx}
\usepackage{textcomp}
\usepackage{xcolor}
\usepackage{url}
\usepackage{booktabs} 
\usepackage{adjustbox}
\usepackage{multirow}
\usepackage{threeparttable}
\usepackage{graphicx} 
\usepackage{tabularx}
\usepackage{hyperref}

\def\BibTeX{{\rm B\kern-.05em{\sc i\kern-.025em b}\kern-.08em
    T\kern-.1667em\lower.7ex\hbox{E}\kern-.125emX}}
\begin{document}

\title{Self-Attention to Operator Learning-based 3D-IC Thermal Simulation}


\author{

\fontsize{8.8pt}{8pt}\selectfont
\vspace{1pt}
Zhen Huang$^{1,2}$, Hong Wang$^{1}$, Wenkai Yang$^{3,5}$, Muxi Tang$^{4}$, Depeng Xie$^{5}$, Ting-Jung Lin$^{2}$, Yu Zhang$^{1}$, Wei W. Xing$^{6,\dagger}$, Lei He$^{7,\dagger}$ \\

$^1$ School of Computer Science and Technology, University of Science and Technology of China, Hefei, China \\
$^2$ Ningbo Institute of Digital Twin, Eastern Institute of Technology, Ningbo, China \\
$^3$ ShanghaiTech University, Shanghai, China \quad $^4$ Tsinghua University, Beijing, China \\
$^5$ BTD Technology, Ningbo, China \quad $^6$ University of Sheffield, Sheffield, UK \\
$^7$ University of California, Los Angeles, USA \\

$^{\dagger}${E-mail: \{w.xing@sheffield.ac.uk, lhe@ucla.edu\}} \\

\thanks{$^\dagger$ Corresponding authors. This work was partially supported by “Science and Technology Innovation in Yongjiang 2035” (2024Z283) and by research support from BTD Inc.}
\thanks{This paper has been accepted by the Design Automation Conference (DAC) 2025.}

\vspace{-20pt}

}

\maketitle

\begin{abstract}

Thermal management in 3D ICs is increasingly challenging due to higher power densities. Traditional PDE-Solving based methods, while accurate, are too slow for iterative design. Machine learning approaches like FNO provide faster alternatives but suffer from high-frequency information loss and high-fidelity data dependency. We introduce \underline{S}elf-\underline{A}ttention \underline{U}-Net \underline{F}ourier \underline{N}eural \underline{O}perator (\textbf{SAU-FNO}), a novel framework combining self-attention and U-Net with FNO to capture long-range dependencies and model local high-frequency features effectively. Transfer learning is employed to fine-tune low-fidelity data, minimizing the need for extensive high-fidelity datasets and speeding up training. Experiments demonstrate that SAU-FNO achieves state-of-the-art thermal prediction accuracy and provides an 842× speedup over traditional FEM methods, making it an efficient tool for advanced 3D IC thermal simulations.
\end{abstract}


\section{Introduction}

Thermal simulation in Electronic Design Automation (EDA) is essential as integrated circuit technology advances. Compared to traditional 2D and 2.5D designs, 3D integration has higher power density, making it more susceptible to overheating and hot spots, which can degrade performance and shorten chip lifespan~\cite{thermal1, thermal2, thermal3, thermal4, wei2025modelgen, wei2025vflow}. Efficient thermal management and rapid simulation tools are therefore vital for layout optimization and reliability~\cite{floorplan1}.

3D chip thermal simulation involves solving the 3D heat conduction PDE. Traditional methods like the Finite Element Method (FEM) and Finite Difference Method (FDM), as used in tools like Ansys Icepak, Celcius, COMSOL, and MTA, are accurate but often too slow due to the complexity of mesh partitioning and PDE solving, limiting their use in iterative design optimization~\cite{FEM1, FEM2, FEM3}.

With the rise of artificial intelligence, ML-based approaches have gained attention in thermal simulation~\cite{ML-2, ML-3, ML-5, ML-6, ML-7, ML-8}. For example,~\cite{CNNthermal} introduced a CNN model for temperature field prediction, but purely data-driven methods lack physical interpretability and need extensive training data.

PINN~\cite{PINN1}, which parameterizes PDE solutions directly through neural networks, has been used for various heat transfer problems~\cite{ePINN, pinn-2}. Its main advantage is embedding physical laws into the learning process for consistent results. However, PINN models require retraining for different system instances due to their fixed geometry, limiting scalability in 3D IC thermal simulations. DeepOHeat~\cite{deepoheat} improved this by combining PINN with operator learning to map power distribution and heat transfer, though at the cost of long training times. Despite leveraging DeepONet’s~\cite{lu2019deeponet} grid-invariance, training duration remains a significant challenge.

FNO uses Fast Fourier Transform (FFT)~\cite{FNO} to approximate PDEs, making it more consistent with physical laws than CNNs. It performs well in the Darcy flow and the Navier-Stokes equation. The resolution-invariant capability ensures FNO's adaptability to different chip sizes, a key advantage in thermal simulations~\cite{FNO1, FNO2, wang2024aro}. Recent work, such as MLA-FNO~\cite{FNO-MRA}, improves precision and speed by combining active learning and FNO. However, FNO methods are limited by two significant challenges. First, its tendency to neglect high-frequency components results in notable errors. Secondly, FNO still demands considerable high-fidelity simulation data for effective model training.

Inspired by U-Net's ability to preserve multi-scale features through skip connections~\cite{UNET, wang2018non}  and self-attention's~\cite{vaswani2017attention} capacity to capture long-range dependencies, we integrated these mechanisms with the FNO for 3D IC thermal modeling. This combination significantly improves the accuracy of thermal predictions, particularly in capturing local temperature variations and high-frequency details, leading to more precise thermal management and design. Additionally, we leverage transfer learning~\cite{wang2024transfer} by fine-tuning the model trained with low-fidelity data, further improving performance and reducing the data requirements. This approach enhances the accuracy of thermal predictions, particularly in capturing local temperature variations and high-frequency details, and addresses the inherent challenge of data scarcity in machine learning applications.
Our key contributions are as follows:
\begin{itemize} 
    \item We propose a novel operator learning architecture,  \underline{\textbf{S}}elf-\underline{\textbf{A}}ttention \underline{\textbf{U}}-Net \underline{\textbf{F}}ourier \underline{\textbf{N}}eural \underline{\textbf{O}}perator (\textbf{SAU-FNO}), that integrates attention mechanisms and U-Net with FNO to enhance 3D IC thermal modeling, particularly in capturing high-frequency information, making it better suited for 3D IC thermal simulations. 
    \item We incorporate transfer learning to fine-tune the model with low-fidelity data, which avoids reliance on high-fidelity data while achieving competitive accuracy. 
    \item Extensive experiments demonstrate that our method achieves state-of-the-art (SOTA) results, with metrics Mean Squared Error (MSE) reduced by over 50\% compared to other approaches. 
    \item SAU-FNO achieves significant computational acceleration, delivering over 842× speedup compared to mainstream commercial software such as COMSOL and MTA, while maintaining comparable accuracy. 
\end{itemize}

The rest of the paper is structured as follows. Section~\ref{sec2} provides the background and preliminaries of thermal modeling and related technology. Section~\ref{sec3} describes the proposed SAU-FNO in detail. Section~\ref{sec.4} presents the experimental results, and Section~\ref{sec5} summarizes the paper.

\section{Background}\label{sec2}

\subsection{Thermal modeling of 3D IC}

The core of 3D IC thermal analysis is simulating heat conduction. Thermal analysis tools compute temperature distribution across the chip based on power routing and key physical parameters like chip layout and stack geometry, enabling accurate heat distribution evaluation and thermal management optimization. The main challenge lies in solving the global heat conduction PDEs:

\begin{equation}
\sum_{i=1}^{3} \frac{\partial}{\partial y_i} \left( k \frac{\partial T}{\partial y_i} \right) + Q_g = \rho c_p \frac{\partial T}{\partial t},
\label{eq:PDE}
\end{equation}
Where $C_p$, $\rho$, and $k$ are the capacity, mass density, and thermal conductivity of the material, respectively. $Q_g$ and $T$ stand for the rate of internally generated energy per unit volume at any spatial-temporal location and the temperature.

For simplicity, we consider a steady temperature field of isotropic materials, simplifying the equation as follows:
\begin{equation}
C_v \frac{d T}{d t} = k \nabla^{2} T + Q_g,
\label{eq: simplified}
\end{equation}
Where $C_v$ represents the volume heat capacity of a substance, $\frac{d T}{d t}$ represents the rate of change of temperature with time, $k$ represents the thermal conductivity, $\nabla^{2} T$ represents the Laplacian operator, and $Q_g$ the rate of production of heat inside the control volume represents the volume heat generation rate.


We consider the steady-state simulation problem
and get the following simplified formula:
\begin{equation}
k \cdot \nabla^2 T + Q_g = 0
\label{eq:simplest}
\end{equation}
the temperature distribution $t(x)$ of 3D IC is governed by the boundary condition, where Robin condition and Neumann condition are adopted:
\begin{equation}
\begin{aligned}
    \kappa(x) \nabla t(x) \cdot n &= g(x) \\
    \kappa(x) \nabla t(x) \cdot n &= \eta(t_{a}-t(x))
\end{aligned}
\end{equation}
where $t(x)$ represents the temperature at position $x$, $\eta$ is the thermal transmissivity. $f(x)$ denotes power trace distribution. 

\subsection{Fourier Neural Operators}
Neural networks excel at learning mappings in finite-dimensional spaces but struggle with parameter or boundary condition variations, limiting generalizability. Neural operators address this by learning mappings between function spaces. The Fourier Neural Operator (FNO) efficiently maps initial conditions to PDE solutions, making it valuable for complex physical problems.

The operator mapping is mathematically described as:
\begin{equation}
\mathcal{G}: \mathcal{U} \rightarrow \mathcal{V},
\end{equation}
where \(\mathcal{G}\) maps inputs from the function space \(\mathcal{U}\) to outputs in \(\mathcal{V}\). This mapping generalizes across various input scenarios within these spaces, not confined to single instances.

FNOs leverage FFT to map PDE solutions into the frequency domain, enhancing generalization and computational speed compared to traditional models like CNNs. Their resolution invariance and scalability make them ideal for fluid mechanics (e.g., Navier-Stokes equations) and efficient for large-scale 3D IC thermal simulations across varying chip sizes and grid resolutions.

The core functionality of FNOs involves transforming the input function \( u_\ell(x) \) through a series of layers, each performing the operation:
\begin{equation}
u_{\ell+1}(x) = \sigma \left( W u_\ell(x) + \mathcal{F}^{-1} \left( \rho(\xi) \cdot \mathcal{F}\left( u_\ell(x) \right) \right) + b \right).
\end{equation}
Here, \( \mathcal{F} \) and \( \mathcal{F}^{-1} \) denote the Fourier and inverse Fourier transforms, respectively. The term \( \rho(\xi) \) represents a learnable convolution kernel in the frequency domain, \( W \) is a linear operator (weight matrix), \( b \) is a bias term, and \( \sigma \) is the Gaussian Error Linear Unit (GELU). 

By using Fourier transforms, FNOs capture global and local patterns critical for modeling complex PDE behaviors, enabling generalization across resolutions and geometries for various physical problems.
In summary, FNOs advance neural network architectures for PDE modeling with their operator-learning capability, resolution invariance, and computational efficiency, making them powerful for complex simulations, including thermal modeling.

\subsection{Self-attention}
Self-attention, first popularized in NLP~\cite{vaswani2017attention}, captures global information and long-distance dependencies more effectively than traditional networks. By dynamically adjusting data relationships through queries, keys, and values, it excels with complex, high-dimensional data. Unlike CNNs focusing on local fields, self-attention learns global interdependencies, enhancing recognition of broad patterns and fine details.



In applications such as 3D IC thermal modeling, self-attention not only tracks regional temperature correlations but also identifies local temperature spikes, adapting seamlessly to large-scale data. This capability enhances detail capture, particularly of high-frequency components, thus boosting both the accuracy and efficiency of simulations.

\subsection{Transfer learning}
Transfer learning transfers knowledge from pre-trained models to solve new tasks, particularly useful when data is scarce or costly~\cite{wang2024transfer}. It leverages features learned from a related task and fine-tunes them for the target task, reducing reliance on large amounts of high-fidelity data while improving model performance and convergence speed.

In 3D IC thermal simulation, obtaining high-fidelity data is expensive and time-consuming. We addressed this by initially training a model with low-fidelity data to learn basic heat transfer patterns and global features quickly. The model was then fine-tuned with a small set of high-fidelity data to capture detailed features, like local hot spots and junction temperatures. This strategy enhances complex temperature distribution prediction, reduces high-fidelity data requirements, and accelerates convergence, offering an efficient solution for large-scale simulations.



\section{Proposed Method}\label{sec3}

Our method \textbf{SAU-FNO} consists of three stages: lifting the input to a higher-dimensional space, passing it through iterative U-Fourier layers, and projecting the result back into the original space. Each U-Fourier layer employs a combination of kernel integral transformations, U-Net and CNN operators, and linear operators, enhancing feature extraction.

Additionally, we introduce a self-attention mechanism to capture long-distance dependencies, which is crucial for tasks like 3D thermal simulations. This setup also supports transfer learning, allowing training on low-resolution data and fine-tuning with high-resolution data for improved generalization.

\subsection{Fundamental U-FNO Design}\label{sec3.1}
\begin{figure}[t]
    \centering
    \includegraphics[width=0.45\textwidth]{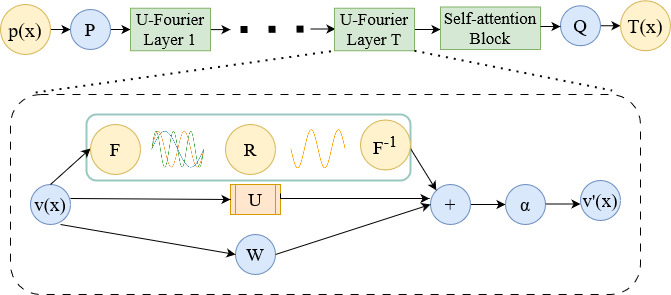}
    \caption{Our SAU-FNO model architecture. $p(x)$ is the input, $P$ and $Q$ are fully connected neural networks, and $T(x)$ is the output. $\mathcal{F}$ denotes the Fourier transform, R is the parameterization in Fourier space, $\mathcal{F}^{-1}$ is the inverse Fourier transform, $W$ is a linear bias term, and $\alpha$ is the activation function. $U$ denotes a U-Net.}
    \label{fig:U-FNO}
\end{figure}

As described in Section 2, FNO consists of multiple superimposed Fourier layers. We enhance the original Fourier layers by integrating a U-Net bypass in each U-FNO layer, as illustrated in Figure~\ref{fig:U-FNO}. The U-Net processes local convolutions, enhancing the representation of high-frequency information.
The specific steps of our proposed method are as follows:
\begin{enumerate}
    \item \textbf{Lifting:} Lift the input observation $a(x)$ to high dimensional space using a neural network transformation $P$.
    \item \textbf{Iterative Layers:} Apply iterative Fourier layers followed by iterative U-Fourier layers:
    \begin{equation}
        v_{l_0}(x) \rightarrow \cdots \rightarrow v_{l_L}(x) \rightarrow v_{m_0}(x) \rightarrow \cdots \rightarrow v_{m_M}(x),
    \end{equation}
    where $v_{l_k}(x)$ for $k=0,1,\ldots,L$ and $v_{m_k}(x)$ for $k=0,1,\ldots,M$ are sequences of functions taking values in $\mathbb{R}^c$, with channel dimension $c$.
    \item \textbf{Projection:} Project $v_{m_M}(x)$ back to the original space using a fully connected neural network transformation $Q$, yielding $z(x) = Q(v_{m_M}(x))$.
\end{enumerate}

Within each newly proposed U-Fourier layer, we define:
\begin{equation}
    v_{m_{k+1}}(x) = \sigma \left( \mathcal{K}v_{m_k}(x) + \mathcal{U}v_{m_k}(x) + W v_{m_k}(x) \right),  \forall x \in D,
\end{equation}
where $\mathcal{K}$ is the kernel integral transformation, $\mathcal{U}$ is a U-Net CNN operator, and $W$ is a linear operator; all are learnable parameters. The activation function $\sigma$ is GELU, which introduces strong non-linearity into each U-Fourier layer. Other settings are detailed in Section~\ref{sec.4}, Experiment Settings.

\subsection{Attention-Enhanced U-FNO}
Attention mechanisms, widely used in computer vision models like Vision Transformer (ViT), excel at capturing key spatial relationships. Similarly, thermal fields exhibit complex spatial temperature variations. By leveraging self-attention, models can capture long-range dependencies, enhancing the understanding of global heat conduction effects. Inspired by this, we integrate a self-attention module to improve feature extraction in 3D IC thermal simulation tasks.

\begin{figure}[htbp]
    \centering
    \includegraphics[width=0.45\textwidth]{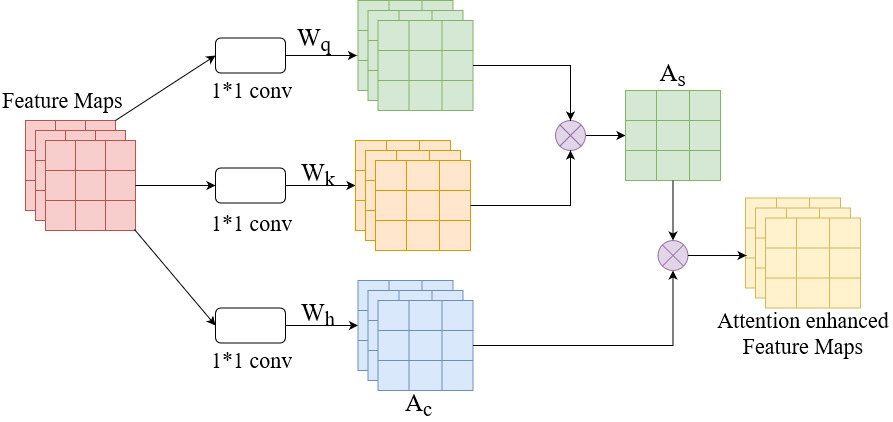}
    \caption{The self-attention block architecture.}
    \label{fig:attention}
\end{figure}

The overall self-attention block architecture is shown in Figure~\ref{fig:attention}. We perform a $1 \times 1$ convolution on the feature map $V_t(x)$ obtained from the U-FNO to compute the spatial attention map $A_s$ and the channel attention map $A_c$ sequentially. The computations are as follows:
\begin{equation}
\begin{aligned}
    A_c &= W_h V_t(x),\quad Q = W_q V_t(x), \\
    K &= W_k V_t(x),\quad s_{ij} = Q_i^\top K_j, \\
    A_s &= \text{softmax}(s_{ij}) = \frac{\exp(s_{ij})}{\sum_{k=1}^{N} \exp(s_{ik})},
\end{aligned}
\end{equation}
where $W_h$, $W_q$, and $W_k$ are learnable parameters representing the embeddings for the value, query, and key, respectively. The softmax function ensures that the attention weights are properly normalized. These parameters are trained jointly with the U-Fourier layer during training.

By using $1 \times 1$ convolutions in the attention block, we retain the mesh-invariant property of the original U-FNO, preserving its ability to generalize from coarse grid training to high-resolution grid inference. This approach also facilitates transfer learning using multi-fidelity data.

The final attention-enhanced feature map is obtained by combining the spatial and channel attention maps:
\begin{equation}
    V_t^\prime(x) = A_s \otimes A_c,
\end{equation}
where $\otimes$ denotes element-wise multiplication.

As described in Section~\ref{sec3.1}, the architecture of our proposed U-FNO layer is:
\begin{equation}
    v_0 \rightarrow v_1 \rightarrow v_2 \rightarrow \cdots \rightarrow v_t,
\end{equation}
which directly affects the final thermal modeling performance. We refine the U-FNO layer by inputting the feature map $V_t(x)$ to obtain the attention-enhanced feature map $V_t^\prime(x)$, as shown in Figure~\ref{fig:attention}.

The experiments show that adding self-attention blocks after all U-FNO layers yields similar performance to adding them only after the last one or two layers. We attribute this to the multi-layer architecture of U-FNO, where features are gradually transmitted and accumulated. Moreover, the U-Net's skip connections facilitate the propagation of key information, even when attention is applied only in the last layer.

Therefore, to reduce computational complexity, we introduce the attention block only in the last layer, in the $V_t(x) \rightarrow V_t^\prime(x)$ step. Refer to Section~\ref{sec.4} for other detailed settings.


\subsection{Training with Transfer Learning}
SAU-FNO effectively models 3D ICs and, with its mesh invariance, can train on low-resolution data and infer on high-resolution data, enabling better generalization than traditional networks. However, large grid resolution differences between training and testing may reduce performance, as low-resolution data often misses complex local and high-frequency details.

Training directly with high-resolution data is costly and impractical. Transfer learning addresses this by pre-training on low-resolution data and fine-tuning with a small amount of high-resolution data, boosting prediction accuracy and reducing reliance on extensive high-fidelity datasets, thus improving efficiency.


\begin{enumerate}
    \item \textbf{Pre-training:} First, a large amount of low-resolution data $D_l = \{X_l, Y_l\}$ is used to train a low-fidelity model $M_l$. Here, $Y_l$ is the prediction obtained by $M_l$ to the input $X_l$. We use the L2 loss function to optimize the parameters of the model:
    \begin{equation}
        L_2 = \frac{1}{N}\sum_{i=1}^{N} \left( T_{\text{pred}, i} - T_{\text{true}, i} \right)^2,
    \label{l2loss}
    \end{equation}
    where $N$ is the number of samples, $T_{\text{true}, i}$ represents the true value for sample $i$, and $T_{\text{pred}, i}$ represents the model's prediction for sample $i$.
    

    \item \textbf{Fine-tuning:} In the second stage, the pre-trained model $M_l$ is fine-tuned using a relatively small amount of high-fidelity data $D_h = \{X_h, Y_h\}$. The training process is similar to the pre-training stage. The learning rate $lr_2$ during fine-tuning is about an order of magnitude smaller than that used in the pre-training stage. The rest of the settings are detailed in the experimental section.
\end{enumerate}


\section{EXPERIMENTAL RESULTS}\label{sec.4}
\subsection{Experiment Settings}
\begin{figure}[htbp] 
    \centering
    \includegraphics[width=0.5\textwidth]{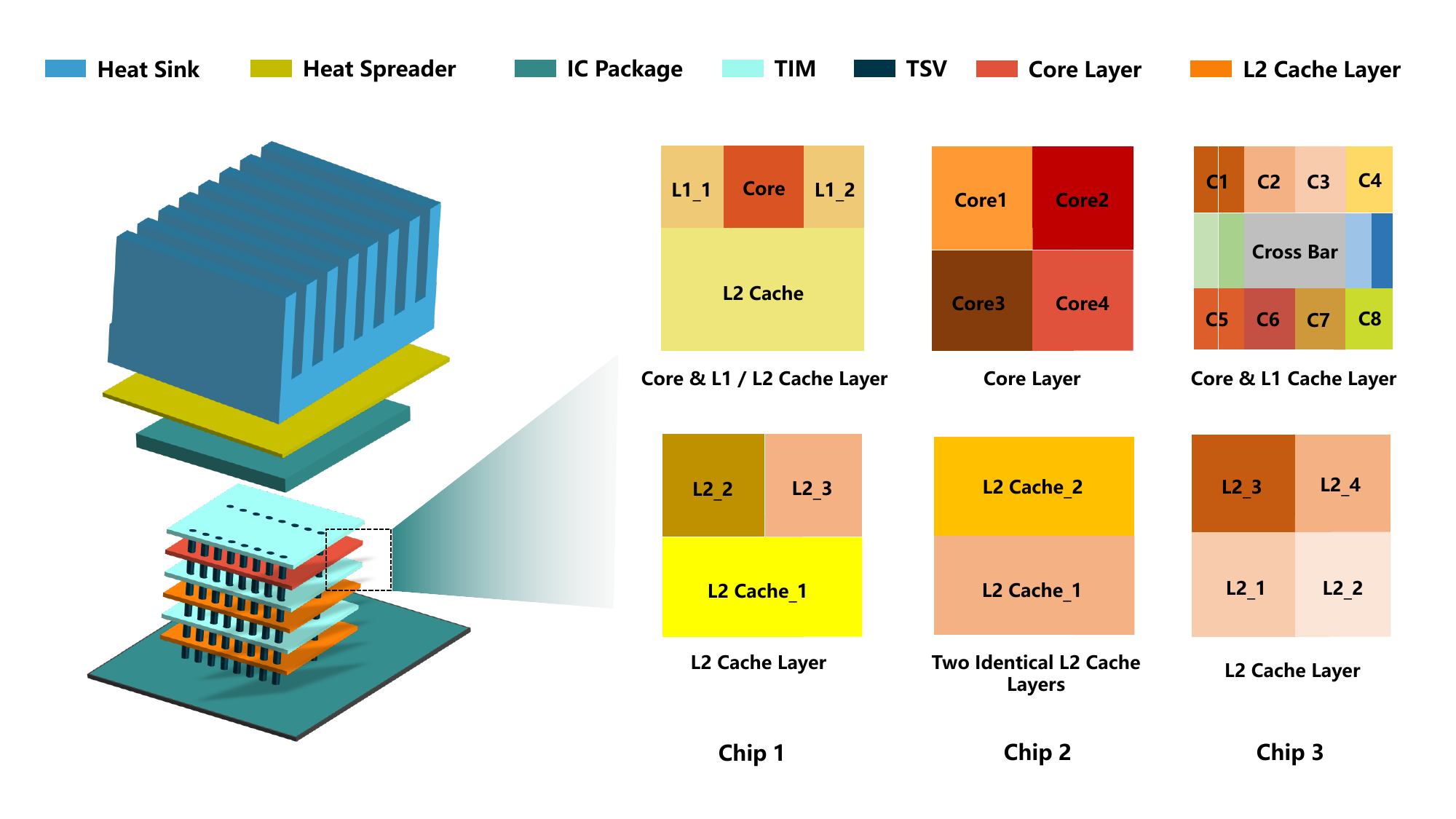} 
    \caption{The stacking structures and floorplan of three 3D chips examined in this work.} 
    \label{fig:architech} 
\end{figure}

\textbf{Experimental Setup}: We adopted the Alpha 21264 EV6 microprocessor architecture~\cite{kessler1999alpha} as the core structure and selected three different 3D ICs for our experiments. These ICs share the same face-to-back stacking structure, including device layers, Thermal Interface Material (TIM), Through-Silicon Vias (TSVs), packaging, and a heatsink. Specifically, Chip1 is a single-core processor consisting of two device layers: one layer contains the core, two L1 caches, and one L2 cache, while the other layer comprises three L2 caches. Chip2 is a three-layer quad-core processor, with the top layer closest to the heatsink consisting of four cores, and the remaining two layers identical, each containing two L2 caches. Chip3 is an octa-core two-layer microprocessor with an architecture similar to Chip1, where the upper layer is composed of eight cores and L1 caches, and the lower layer consists of four L2 caches. In all chips, the address and data bus between the L2 cache and the processor core are connected via TSVs. The 3D structure is illustrated in Figure~\ref{fig:architech}. To simplify the figure of the floorplan, TSVs passing through each layer have been omitted. Additionally, Table~\ref{table:para} provides detailed structures of the 3D ICs.

\textbf{Data Generation}: We randomly assigned power levels to different functional blocks while ensuring the total power remained within an appropriate range. By varying the power distribution, we aimed to achieve significant and reasonable temperature differences in the simulation results, reflecting real-world scenarios. The model output is a three-dimensional temperature distribution; due to the slow processing speed of commercial software such as Celsius, Icepak, and COMSOL, we adopted MTA~\cite{MTA} as the solver to derive the training data outputs. For each of the three types of chips, we generated 5,000 sets of distinct power distributions. Unless otherwise specified, the data was divided into training and testing sets at a ratio of 4:1.

\begin{table*}[htbp]
    \centering
    \setlength{\abovecaptionskip}{0cm}
    \caption{Geometric Structures and Thermal Parameters of 3D-ICs}
    \vspace*{-0.03in}
    \label{table:para}
    \begin{tabular}{c|c|c|c|c|c}
    \toprule[1.5pt]
        & \multicolumn{3}{c|}{size (mm)} & \multicolumn{2}{c}{Thermal Attributes}  \\ \hline
        & Single-Core & Quad-Core & Octa-Core & \begin{tabular}[c]{@{}c@{}}Thermal Conductivity\\ (W/m$\cdot$K)\end{tabular} & \begin{tabular}[c]{@{}c@{}}Volumetric Heat Capacity\\ (J/m$^3$K)\end{tabular} \\ \hline
        L2 Cache Layers & 16$\times$16$\times$0.15 & 12.4$\times$12.76$\times$0.15 & 10$\times$10$\times$0.1 & 100 & 1.75$\times$10$^6$ \\ \hline
        Core layer & 16$\times$16$\times$0.15 & 12.4$\times$12.76$\times$0.15 & 10$\times$10$\times$0.1 & 100 & 1.75$\times$10$^6$ \\ \hline
        TIM & 16$\times$16$\times$0.02 & 12.4$\times$12.76$\times$0.02 & 10$\times$10$\times$0.052 & 4 & 4.00$\times$10$^6$ \\ \hline
        Heat Spreader & 30$\times$30$\times$1 & 30$\times$30$\times$1 & 30$\times$30$\times$1 & 400 & 3.55$\times$10$^6$ \\ \hline
        Heat Sink (Base) & 60$\times$60$\times$6.9 & 60$\times$60$\times$6.9 & 60$\times$60$\times$6.9 & 400 & 3.55$\times$10$^6$ \\ \hline
        Heat Sink (Fins$\times$21) & 1$\times$60$\times$50 & 1$\times$60$\times$50 & 1$\times$60$\times$50 & 400 & 3.55$\times$10$^6$ \\  \hline
        \multicolumn{1}{c|}{TSV} & \multicolumn{3}{c|}{diameter:0.01 ; pitch:0.01} & 100 & 1.75$\times$10$^6$ \\ \hline
        \bottomrule[1.5pt]
    \end{tabular}
\end{table*}

\textbf{Model Setting}: We referenced the settings from previous work and used Optuna and W\&B for hyperparameter tuning. For Chip1 and Chip2, the SAU-FNO was configured with a width of 2 and a model structure of [12, 12, 2]. For Chip3, the width was doubled to 64, resulting in a model structure of [24, 24, 2], to accommodate its larger data scale and grid size. The U-Net encoder has 4 layers of 3×3 convolution with ReLU activations, producing feature maps of [64, 128, 256, 512], while the decoder restores them to [512, 256, 128, 64] using bilinear interpolation and 3×3 convolutions. The self-attention module processes inputs of shape [Bs, 64, 64, 64] into Q, K, and V matrices with dimensions [Bs, 64, 64, d] (d=64). Further settings follow ~\cite{UFNO} and ~\cite{atten1}.

\textbf{Training and Testing}: We utilized a decaying learning rate with the Adam optimizer. The initial learning rate was set to 1e-4, and a weight decay of 1e-5 was applied to mitigate model overfitting. For fine-tuning, an initial learning rate of 1e-5 was used to ensure training stability. Each training process ran for over 200 epochs to guarantee convergence. All experiments were conducted on an RTX 3090 GPU with CUDA 11.3, using PyTorch version 1.10.1 and Python 3.7.

\subsection{Compared to Latest ML methods}
Recently, numerous studies have applied machine learning to 3D IC thermal modeling, but these methods have various limitations. Our approach, as a neural operator method, offers unique properties like mesh invariance not found in traditional neural networks. Direct comparisons of accuracy or time are not entirely fair, so we only benchmarked against recent neural operator methods, including FNO, U-FNO, and DeepOHeat.

\begin{figure}[h!]
    \centering
    \includegraphics[width=0.45\textwidth]{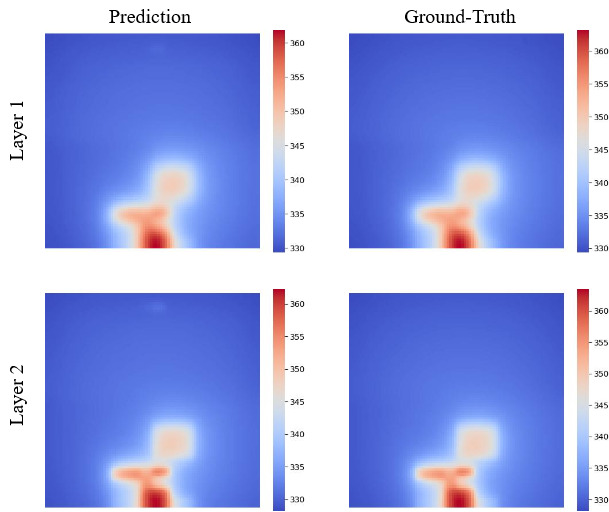}
    \caption{Comparisons between our SAU-FNO predictions and Ground-Truth on Case 1 of Chip1.}
    \label{fig:predict1}
\end{figure}

\begin{figure}[h!]
    \centering
    \includegraphics[width=0.45\textwidth]{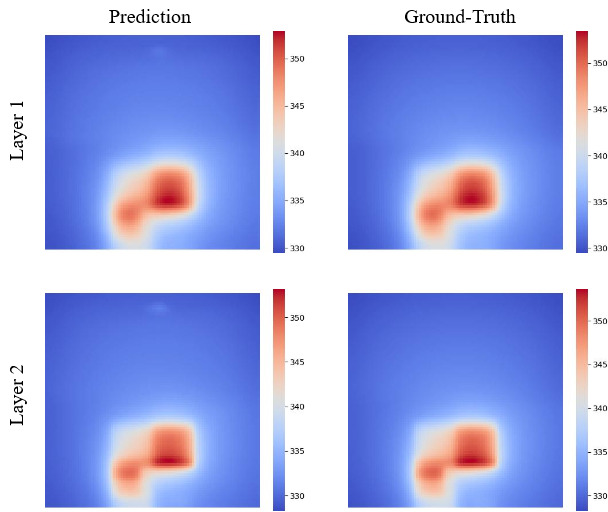}
    \caption{Comparisons between our SAU-FNO predictions and Ground-Truth on Case 2 of Chip1.}
    \label{fig:predict2}
\end{figure}
We selected two representative cases with significant power distribution variations from Chip1 for visualization, as shown in the heatmaps Figure~\ref{fig:predict1} and Figure~\ref{fig:predict2}. The chips have two heating layers, labeled as layer 1 and layer 2. The visualizations show that our thermal simulation results closely match the ground truth, demonstrating strong performance in average temperature difference and excelling in junction temperature.

For quantitative evaluation, we used metrics such as RMSE, average temperature error (Mean), and maximum temperature error (Max), along with MAPE and PAPE. As shown in Table~\ref{table: baseline}, SAU-FNO significantly outperforms the commonly used DeepOHeat and FNO, with RMSE and junction temperature reductions of over 50\% compared to FNO. While methods like U-Net lack resolution invariance and were not extensively compared for fairness, our results demonstrate clear advantages, including significant improvements over GAR, highlighting the superior performance and robustness of our approach. Additionally, the comparison between our method and U-FNO demonstrates the effectiveness of the self-attention module, while the improvements from FNO to U-FNO further highlight the contribution of the U-Net architecture. These comparisons effectively serve as an ablation study validating the importance of each component in our method.

\begin{table}[htbp]
\caption{Comparison of our method with other ML methods, Max stands for junction temperature and Mean stands for average temperature. }
\vspace*{-0.08in}
\label{table: baseline}
\centering
\adjustbox{width=0.5\textwidth, height=!}{%
\begin{tabular}{c|c|c|c|c|c|c}
\hline
\textbf{Method}  & \textbf{Resolution} & \textbf{RMSE}  & \textbf{MAPE}  & \textbf{PAPE} & \textbf{Max}  & \textbf{Mean}  \\
\hline
\multirow{2}{*}{DeepOHeat}\cite{deepoheat} & 40$*$40   &  0.457        & 0.093     & 0.811     & 2.936     & 0.297     \\
                           & 64$*$64  & 0.473        & 0.097     & 0.914     & 4.355     & 0.312     \\
\hline
\multirow{2}{*}{FNO}\cite{FNO}       & 40$*$40   &  0.438     &  0.086  & 0.730   &   2.774  &  0.329    \\
                           & 64$*$64  & 0.409       & 0.077     & 1.073     & 4.081     & 0.292     \\
\hline
\multirow{2}{*}{U-FNO}\cite{UFNO}     & 40$*$40   & 0.221         & 0.049     & 0.195   & 0.741     & 0.185    \\
                           & 64$*$64  & 0.278          & 0.065     & 0.275     & 1.046     & 0.253     \\
\hline
\multirow{2}{*}{GAR}\cite{wang2022gar}       & 40$*$40   & 0.576       &  0.127 & 0.893  &  4.639  & 0.153     \\
                           & 64$*$64  & 0.598         &  0.145    & 0.917    &   5.132   & 0.170     \\
\hline
\multirow{2}{*}{Ours}      & 40$*$40   & 0.197         & 0.041     & 0.168     & 0.650     & 0.146     \\
                           & 64$*$64  & 0.203          & 0.043     & 0.179     & 0.773     & 0.162     \\
\hline
\end{tabular}
}
\end{table}

We found that the structural differences among chips lead to variations in absolute accuracy. However, our method consistently achieves significant improvements and reaches SOTA performance across all chips. Therefore, we only report the average quantitative results for Chip2, with results for other chips provided in the supplementary material.

\subsection{Validation of Transfer Learning}

We implemented transfer learning by pre-training the model on extensive low-fidelity data and fine-tuning it with a small amount of high-fidelity data. This approach achieves nearly the same effectiveness as training from scratch with large-scale high-fidelity data, while significantly reducing the need for such data and accelerating training.

As shown in Table~\ref{table:transfer}, RMSE and junction temperature metrics showed minimal accuracy loss with fine-tuning compared to full high-fidelity training. The low-fidelity model used 4,000 cases, while fine-tuning the high-fidelity model required only 1,000 cases, achieving a 4:1 ratio of low-resolution to high-resolution data, which we determined to be optimal after multiple experiments. A high-fidelity model trained from scratch with 4,000 cases served as a benchmark. Given that high-resolution data collection is 4 to 6 times slower, total training time was reduced by about 2.5 times. Increasing fine-tuning data to 4,000 cases slightly improved accuracy. Transfer learning was also effective on other ML methods like FNO and U-FNO, proving its suitability for efficiency-critical 3D IC thermal simulations.

\begin{table}[htbp]
\centering
\caption{The comparison before and after Transfer learning on chip1, transfer means transfer learning.}
\vspace*{-0.07in}
\label{table:transfer}
\begin{tabular}{c|c|c|c|c|c|c}
\hline
\textbf{Method}  & \textbf{Transfer} & \textbf{RMSE}  & \textbf{MAPE}  & \textbf{PAPE} & \textbf{Max}  & \textbf{Mean}  \\
\hline
\multirow{2}{*}{FNO}\cite{FNO}       & -  &  0.177    &  0.026  & 0.581   & 2.140  &  0.089    \\
                           & \checkmark  & 0.185     & 0.031     & 0.615     & 2.245     & 0.097     \\
\hline
\multirow{2}{*}{U-FNO}\cite{UFNO}     & -   & 0.130     & 0.018     & 0.298   & 1.115     & 0.061    \\
                           & \checkmark  & 0.141     & 0.023     & 0.309     & 1.224     & 0.073     \\
\hline
\multirow{2}{*}{Ours}      & -   & 0.090     & 0.014     & 0.232     & 0.871     & 0.049     \\
                           & \checkmark  & 0.097     & 0.015     & 0.241     & 0.897     & 0.058     \\
\hline
\end{tabular}
\end{table}
\vspace*{-0.1in}
\subsection{Compared to Professional Thermal Simulation Tool}
We compared the accuracy and speed of our model with mainstream 3D IC thermal simulation tools, including commercial software like COMSOL and open-source tools such as MTA~\cite{MTA} and Hotspot~\cite{huang2006hotspot}. COMSOL, based on FEM, is suited for complex engineering, while MTA is an academic FEM tool, and Hotspot uses fast thermal network analysis based on FDM and empirical models.
For simplicity, we only compared junction temperature and minimum temperature here, while other metrics such as MAPE are provided in the supplementary material.
To achieve a certain degree of fairness in comparison, we conducted simulation experiments with them using the finest mesh and maximum resolution available.

Using the method in Section~\ref{sec.4}, we generated 20 power distributions for chip1-chip3 high-fidelity data and ran simulations to measure the maximum temperature and distribution, comparing average errors with our approach. Junction temperature was the primary comparison metric. Table~\ref{table:commercial} shows that SAU-FNO predictions are close to baseline methods, with a maximum error within 0.25K, proving its practical usability. While COMSOL is slower, MTA and Hotspot performed faster. Our SAU-FNO method averaged 0.27s per prediction, compared to 227.31s for MTA and 98.47s for Hotspot, achieving speedups of 842x and 365x, respectively.

\begin{table}[htbp]
\renewcommand{\arraystretch}{1.2}
\small
\centering
\setlength{\abovecaptionskip}{0.2cm} %
\setlength{\belowcaptionskip}{-0.1cm}
\caption{Maximum and minimum temperature comparisons among COMSOL, MTA, Hotspot and SAU-FNO on five steady-state samples.}
\vspace*{-0.07in}
\label{table:commercial}
\begin{threeparttable}
\adjustbox{width=0.5\textwidth, height=!}{%
\begin{tabular}{c|cccccc}
\hline
 &  & COMSOL & MTA & Hotspot & Ours & Error\tnote{*}
\\ \hline
\multirow{2}{*}{$chip_1$} & Max(K)
& 381.257 & 381.173 
& 391.051 & 381.159 
& -0.098 \\
 & Min(K) 
 & 328.032 & 328.107 
 & 334.970 & 328.139
 & +0.107
\\ \hline
\multirow{2}{*}{$chip_2$} & Max(K)
& 380.065 & 380.134 
& 392.015 & 380.195 
& +0.130 \\
 & Min(K) 
 & 325.259 & 325.450 
 & 336.290 & 325.473
 & +0.214
\\ \hline
\multirow{2}{*}{$chip_3$} & Max(K)
& 422.207 & 422.079 
& 432.017 & 422.091 
& -0.108 \\
 & Min(K) 
 & 347.496 & 347.330 
 & 348.019 & 347.353
 & -0.143
\\ \hline
\end{tabular}
}
\begin{tablenotes}
        \footnotesize
        \item[*]Error refers to the difference between SAU-FNO and COMSOL.
        \end{tablenotes}
    \end{threeparttable}
\end{table}


\section{Conclusion}\label{sec5}
We present SAU-FNO, an advanced thermal modeling framework that addresses key challenges in 3D IC thermal management. By integrating self-attention and U-Net architectures into the FNO, SAU-FNO effectively captures both global and local thermal dynamics. Our results demonstrate significant accuracy improvements and an 842× reduction in computation time compared to traditional methods. Transfer learning further boosts performance, reducing reliance on extensive high-fidelity datasets. Future work will expand SAU-FNO's capabilities to support a broader range of thermal analysis tasks, promoting more efficient design processes in advanced semiconductor technologies.

\clearpage

\bibliographystyle{IEEEtran}
\bibliography{sample-base}

\end{document}